\documentclass[pdflatex,sn-mathphys-num]{sn-jnl}

\usepackage{graphicx}%
\usepackage{multirow}%
\usepackage{amsmath,amssymb,amsfonts}%
\usepackage{amsthm}%
\usepackage{mathrsfs}%
\usepackage[title]{appendix}%
\usepackage{xcolor}%
\usepackage{textcomp}%
\usepackage{manyfoot}%
\usepackage{booktabs}%
\usepackage{algorithm}%
\usepackage{algorithmicx}%
\usepackage{algpseudocode}%
\usepackage{listings}%

\theoremstyle{thmstyleone}%
\theoremstyle{thmstyletwo}%

\theoremstyle{thmstylethree}%

\begin{document}

\title[Article Title]{Jointly Predicting Courses and Grades Using a Transformer-Based Model}


\author*{\fnm{Paul} \sur{Savala}}\email{psavala@stedwards.edu}

\affil{\orgdiv{Mathematics}, \orgname{St. Edward's University}, \orgaddress{\street{3001 S Congress Ave}, \city{Austin}, \postcode{78704}, \state{TX}, \country{USA}}}


\abstract{Existing predictive models in learning analytics often treat student academic history as a simple sequence, overlooking the concurrent nature of courses taken within a semester. This simplification can lead to inaccurate performance predictions, particularly for students with heavy or challenging course loads. This paper introduces a \textbf{TR}ansformer for \textbf{A}cademic \textbf{C}ourse-grade \textbf{E}stimation (TRACE) that addresses this limitation by jointly predicting both the set of courses a student will take and their corresponding grades for an upcoming semester. Our approach encodes courses on a per-semester basis to capture the effects of course concurrency and utilizes a novel loss function combining course-set prediction with grade prediction. We demonstrate that predicting courses taken in addition to the grades in those courses leads to significant improvements in prediction quality. Trained on ten years of institutional data, our joint prediction model reduces mean absolute error by nearly 50\% compared to an identical architecture that predicts grades alone. The model also outperforms traditional LSTM-based sequential models, as well as graph neural network-based approaches, and offers natural ways to incorporate student attribute data. This work demonstrates the utility of modern neural architectures for creating interpretable models that can be adapted to new institutions via retraining and recalibration, as well as the importance of key techniques, such as predicting courses taken during training. We discuss how this model could be incorporated into early detection systems at institutions of higher education.}




\maketitle


\section{Introduction}\label{sec1}

\subsection{Predictive models in LA}
The field of learning analytics (LA) has long been interested in the ability to predict course outcomes \cite{shafiq2022student}. In particular, past work has shown that, while grades in prior semesters are strong predictors of success in future semesters \cite{borsato2013college} \cite{balfanz2007preventing}, there are also a number of other mediating factors, such as academic ability, motivation, classroom environment, home environment, television viewing, gender, and race \cite{welch1986predicting}.

Clearly, there is a complex interrelationship between a student's chosen major, their course selection, and their grades in those courses in relationship to their future performance. Two students with different majors and with distinct course histories may perform very differently in the same class. In addition, temporal factors come into play. In particular, past work has shown that the temporal proximity between a prerequisite course and the follow-up has a major effect on the student's performance \cite{dills2016knowledge} \cite{belanger2019class}. However, as students progress throughout their career, the knowledge and expertise needed to perform well in one course may depend on several others. For example, a computer science student taking an algorithms and data structures course may need to draw on their knowledge of discrete mathematics, as well as object-oriented programming.

In addition, course relationships are often highly complex \cite{quitadamo2007learning}. For example, a student who performed well in a writing course may find writing chemistry lab reports simple, but may struggle with the algebra skills needed to calculate the chemical interactions. Indeed, research on students in general education biology course has shown the effects of writing skills in a laboratory setting \cite{quitadamo2007learning}. While simple heuristics can be devised to model certain relationships (e.g. indicating when one course is a prerequisite for another), these may fail to capture the more complex relationships that affect student performance. Moreover, most students take multiple courses in any given semester. This course load is itself a variable and may affect student performance. In addition, the precise courses taken at the same time can be an influencing factor, such as for students taking multiple difficult courses simultaneously.

Besides related courses, students also need to consider the effect of course load. Students select courses according to a number of factors, such as avoiding courses that are perceived to be too difficult, learning value, and lecturer style, among other questions \cite{babad2003experimental}. In addition, the concept of ``Academic Momentum'' suggests that higher course loads earlier in their college career can strongly influence later outcomes \cite{attewell2012academic}. Because of these factors, when modeling student performance and course selection, it's important to consider per-semester course load as factor, rather than treating each course as an independent event.

While prior studies such as \cite{ren2019grade} and \cite{morsy2020context} have incorporated concurrently taken courses through pairwise interactions or attentional modules, our approach differs by leveraging positional encodings where all courses taken in the same semester receive the same temporal vector. This design enforces a permutation-invariant representation within each semester, which we argue more accurately reflects the unordered nature of concurrent enrollments.

\subsection{Research questions}
This study is guided by the following research questions:
\begin{itemize}
\item \textbf{RQ1:} To what extent does a Transformer model that encodes semester-level concurrency improve grade prediction accuracy compared to models that treat course history as a flat sequence?
\item \textbf{RQ2:} Does the joint prediction of courses and grades lead to a significant improvement in grade prediction accuracy compared to an architecture that predicts grades alone?
\end{itemize}

While Transformer architectures are well-established, their application to academic prediction requires resolving several non-trivial design choices that have not been addressed in prior work. Specifically: 
\begin{itemize}
    \item How should concurrent courses be represented in sequence models? Should the representation be as ordered sequences or unordered sets?
    \item Should course identity and grade be predicted jointly or separately, and how does this choice affect learned representations?
    \item What loss function appropriately handles set-valued predictions where order is arbitrary, while not adversely affecting training or inference time? In addition, in the learning analytics context, while order is arbitrary, the pairing of courses and grades is important, so order is only arbitrary up to (course, grade) pair. How should this be handled?
\end{itemize}

These implementation decisions substantially affect model performance, as we demonstrate empirically, yet have received little attention in the learning analytics literature.

Prior work in course and grade prediction has typically treated these as separate tasks. Recent models such as CourseBEACON \cite{khan2023session} and PLAN-BERT \cite{shao2021degree} use sequential or masked language models for course planning or recommendation, but do not jointly predict course sets and associated performance. Our approach addresses this by formulating a joint prediction task that combines course set selection and grade regression in a single multi-task Transformer model. This architectural framing enables TRACE to capture the mutual dependency between which courses a student takes and how well they perform in them, which has not been done in prior work.

In addition, recent work in learning analytics has increasingly framed student performance prediction as a relational learning problem and has adopted graph neural networks (GNNs) \cite{scarselli2008graph} to model complex dependencies among students, courses, and learning behaviors. In this literature, students are typically represented as nodes in a graph, with edges encoding similarity, interaction, or shared attributes. Nodes are then aggregated, and and node embeddings updated, all according to various architectural approaches. Graph convolutional networks and graph attention architectures, among other approaches \cite{zhang2020deep, zhou2020graph}, are then used to propagate information across these relational structures to predict academic outcomes.

Several studies construct student-to-student graphs based on demographic similarity, historical grades, or learning behaviors and demonstrate improved predictive accuracy over tabular or sequence-based baselines. For example, \cite{yu2022graph} model student similarity using a graph neural network to predict academic performance, while \cite{li2020peer} leverage peer interaction graphs derived from online learning environments to capture peer influence effects. Others use multiple graphs, such as \cite{huang2024improving} which models student interactions and student attributes as two separate graphs, and combines them to make predictions. These approaches highlight the effectiveness of relational inductive biases in modeling student performance, and the flexibility of graph neural network approaches.

We make three contributions: 
\begin{enumerate}
    \item we formulate next-semester academic prediction as a \emph{joint} course-set and grade prediction problem in a single model; 
    \item we introduce a semester-level concurrency encoding that is permutation-invariant within a term while preserving temporal order across terms; and 
    \item we propose a set-based multi-task objective (KL for course-set prediction + MSE for grade regression) that avoids order artifacts induced by token-level cross entropy.
\end{enumerate}

In our study we utilize a \textbf{TR}ansformer for \textbf{A}cademic \textbf{C}ourse-grade \textbf{E}stimation (TRACE) to predict grade performance based on student major, course history and grade history. In particular we implement two novel approaches, which we show significantly improve grade predictions.

First, we show that simultaneously predicting courses and grades, using an embedding layer to represent courses, leads to significantly higher performance than a simpler model. Indeed, we experiment with a model which still uses course embeddings, but only predicts grades, and note a mean absolute error nearly double that of the model which jointly predicts both, and a mean squared error 3.5 times higher. We also show that predicting just grades, but not courses, leads to significantly reduced performance. Therefore, both factors (simultaneous course and grade predictions, combined with course embeddings) are extremely important. 

Despite this, the purpose of joint course prediction is \textit{not} to recommend courses to students, but rather to force the model to learn meaningful course representations. This is analogous to how auxiliary tasks in multi-task learning improve representation quality for the primary task. This is analogous to word embeddings such as those trained by BERT. The primary task was masked word prediction (predicting a missing word based on the surrounding context). However, the training phase had a secondary task of predicting the next sentence, as this leads to stronger sentence relationships, and results in better downstream prediction quality \cite{devlin2019bert}.

Second, we maintain the concurrency of courses taken with the semester by utilizing positional encoding, using this to predict all courses in the subsequent semester simultaneously. Doing so allows the model to better understand the relative temporal relationship between courses, but also to predict an entire semester's worth of courses simultaneously. For contrast, \cite{polyzou2016grade}, \cite{okubo2017students}, \cite{sweeney2015next} and \cite{nachouki2023student} all use course history to predict future grades, but treat all courses as occurring sequentially, regardless of the semester in which the course was taken. Our approach requires us to introduce a custom loss function which supports this concurrency.

As discussed in the comprehensive review in \cite{cui2019predictive}, the majority of the research used course-specific models to predict performance. They note that ``it is worth investigating whether a general prediction model can be developed for use in multiple courses.'' Our work is not unique in addressing this problem, but contributes to the literature of general purpose models which can be broadly deployed and easily updated when new courses are implemented.


\section{Methods}\label{sec3}
This study was conducted at an American university, in the context of a LA project which aimed at modeling course performance based on historical course selection, grades and student major. The university collects data on student registration and course grades, and this data was combined with data on student majors. This study obtained anonymized study data consisting of over 5000 students across ten years. All courses were offered in either the fall or spring semester. We sought and received IRB approval for use of this data from the university's IRB. The university's informational technology division provided the data.

\subsection{Dataset}
The dataset consisted of 5326 students enrolled in 360 different courses across 48 majors and 18 semesters. Data extended from the Fall 2014 semester through the Fall 2023 semester, and included all fall and spring semesters in this date range. Students in the data were split into training and testing sets, consisting of 90\% and 10\% of all students, respectively. 

We take the full academic history of the student and create multiple (input, output) pairs from their academic history. We make the first input be the student's major, courses and grades from semester 1, and the output be the courses and grades from semester 2. Then, assuming the student has at least three semesters of academic history, we make the next input be the courses and grades from semesters 1 and 2, and the output be the courses and grades from semester 3. We repeat this process for the students entire academic history and for all students. 

\subsection{Model Architecture}
\begin{figure}[h]
    \includegraphics[width=\linewidth]{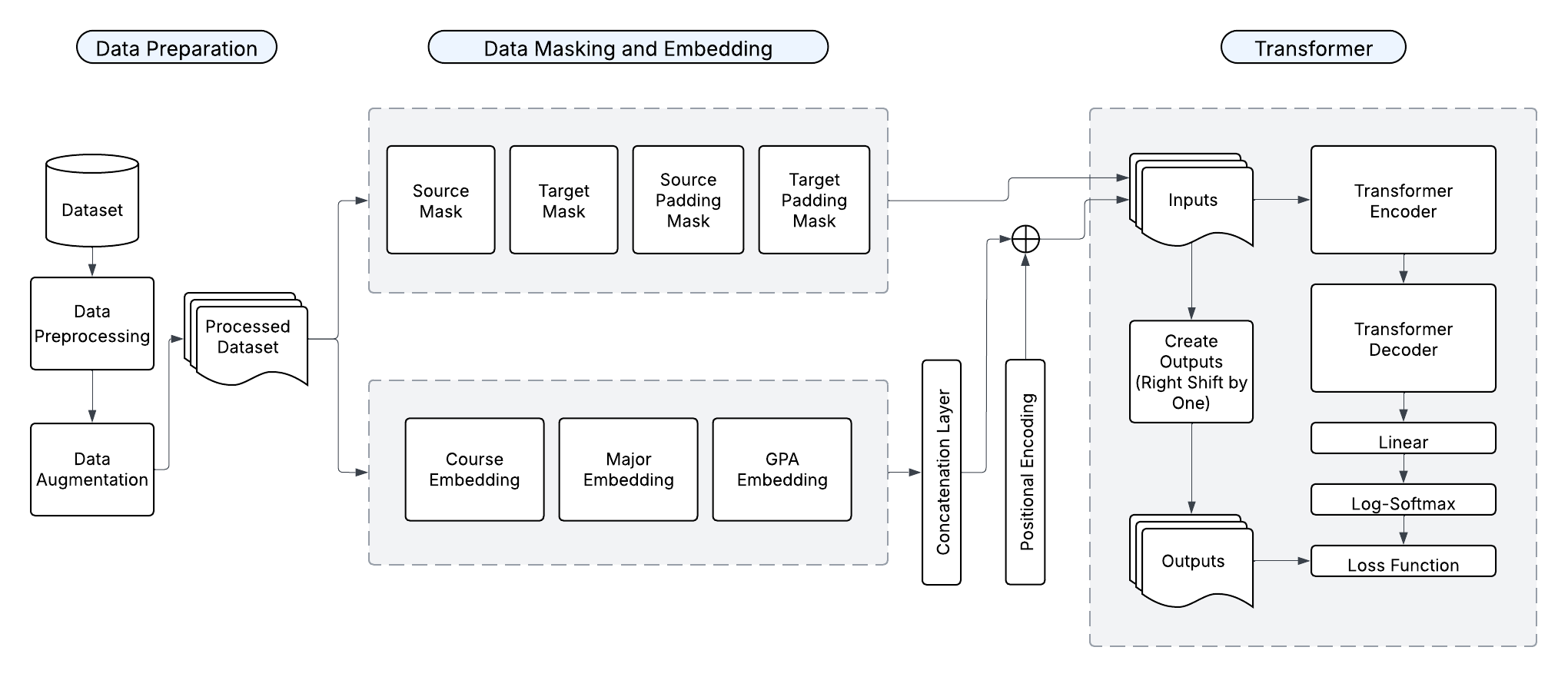}
    \label{fig:architecture}
    \caption{Overview of model architecture}
\end{figure}

In this section we detail our data preparation, and give an overview of the TRACE model architecture. Transformer models were originally developed in the context of machine translation, and thus much of the work around them is described in a natural language processing setting. In this section we give examples showing how there is a natural relationship between language modeling and course and grade prediction.

\subsubsection{Data Preparation}
Data preparation consisted of data cleaning and standardization. We filtered out students who had completed less than ten courses at the university, as well as those who had taken more than sixty courses. Doing so removed students with insufficient course history for prediction, or who had a course history much longer than is typically expected, perhaps indicating a situation which may not generalize. We label encoded all course names to sequential integers in a random order, ignoring the effect of course subject or number. This was done purposefully to allow the model to learn latent representations of these courses, without relying on hand-encoded features. 

\subsubsection{Data Augmentation}
Courses that had been taken less than 100 times in total across the entire dataset were all encoded as an \texttt{<OTHER>} token. Similarly, majors were label encoded, and majors with less than five hundred students ever having declared them were encoded as an \texttt{<OTHER>} token. Grades were converted to standardized GPA values between 0 (F) and 4.3 (A+). 

Sequence models are typically standardized \cite{wu2016google} in the following way: All sequences are prefixed with a \texttt{<SOS>} token to indicate the start of the sequence, appended with a \texttt{<EOS>} token to indicate the end of the sequence, and then padded to a common length using \texttt{<PAD>} token. Since Transformer models process inputs in fixed-size batches, shorter sequences are typically padded to match the length of the longest sequence in a batch. We do the same for our data, padding all students to have semesters with the same length, where padding is appended to enforce this shared length.

\subsubsection{Data Masking}
Masking is a fundamental technique employed in Transformer architectures to control information flow and maintain appropriate contextual boundaries. The self-attention mechanism inherently connects all positions to each other, necessitating strategic constraints through various masking approaches to preserve model integrity and enable specific functionalities \cite{vaswani2017attention}.

Padding masks (also called attention masks) address the variable-length nature of input sequences. To prevent the model from attending to these artificial padding tokens, a padding mask is applied:

\begin{align}
\text{Mask}_{\text{padding}}(i, j) = 
\begin{cases}
0 & \text{if position } j \text{ contains a padding token} \\
1 & \text{otherwise}
\end{cases}
\end{align}

This binary mask is applied to the attention scores, effectively setting attention scores for padding tokens to $-\infty$, resulting in zero attention weight after softmax \cite{vaswani2017attention, liu2019roberta}.

In autoregressive decoding scenarios, such as those employed in the Transformer decoder, each token should only attend to previous tokens in the sequence to prevent information leakage from future positions during training. For example, we would not want our model to know how a student performed in Calculus 2 when predicting their grade in Calculus 1. This is achieved through causal masking (sometimes called target masking or look-ahead masking):

\begin{align}
\text{Mask}_{\text{causal}}(i, j) = 
\begin{cases}
0 & \text{if } j > i \\
1 & \text{if } j \leq i
\end{cases}
\end{align}

This triangular mask ensures that position $i$ can only attend to positions $j \leq i$, preserving the autoregressive property \cite{vaswani2017attention}. In our situation, we utilize attention masks to ensure that the model attends only to previous semesters.

These masking mechanisms collectively enable Transformers to maintain sequence ordering constraints, handle variable-length inputs efficiently, and implement diverse training objectives, contributing significantly to the architecture's versatility across a wide range of sequence modeling tasks.

\subsubsection{Embedding Layers}
Transformer models begin by converting discrete input tokens into continuous vector representations through embedding layers. These learned embeddings map tokens from a vocabulary to dense vectors in a high-dimensional space where semantic relationships can be captured \cite{mikolov2013distributed}.

In the original Transformer architecture, separate embedding layers are used for both the encoder and decoder inputs \cite{vaswani2017attention}. The embedding dimension is typically set to $d_{model}$ (e.g., 512), which remains consistent throughout the model. Notably, Vaswani et al. multiply these embeddings by $\sqrt{d_{model}}$ to scale the initial embeddings:

\begin{align}
\text{Embedding}(x) = \text{Emb}(x) \cdot \sqrt{d_{model}}
\end{align}

This scaling helps maintain appropriate magnitude before adding positional encodings and entering the encoder/decoder stacks \cite{vaswani2017attention}.

In our work we create separate embedding layers for each major, course, and grade. These embeddings are all independent and learned jointly, allowing the model to learn and leverage the interaction between these factors. While the grade is already numeric, it is well known that grades are not assigned linearly, and grade distributions often depend on the level of the course \cite{sonner2000adjunct}. Therefore, passing grades through a small embedding layer allows the model to capture this nonlinearity.

\subsubsection{Positional Encoding}
In Transformer models, unlike recurrent neural networks, there is no inherent sequential information processing. However, sequence order is crucial for many natural language processing tasks. To address this limitation, positional encoding is introduced to inject information about the position of tokens in the sequence.

The standard approach, as proposed by \cite{vaswani2017attention}, utilizes sinusoidal functions to create position-dependent representation vectors:

\begin{align*}
PE_{(pos,2i)} &= \sin\bigl(pos/10000^{2i/d_{model}}\bigr)\\
PE_{(pos,2i+1)} &= \cos\bigl(pos/10000^{2i/d_{model}}\bigr)
\end{align*}

where $pos$ represents the position in the sequence, $i$ represents the dimension, and $d_{model}$ is the dimensionality of the model. This formulation allows the model to attend to relative positions, as the positional encoding for a fixed offset can be represented as a linear function of the positional encoding at the original position \cite{vaswani2017attention}.

In our situation, the position is represented by the relative semester in which the student took the course. That is, semesters are encoded as 1 (first semester), 2 (second semester), etc. This allows the model to learn the impact of different temporal lags. Because students typically take multiple courses in a single semester, multiple courses will share the same positional encoding. We use relative semesters rather than absolute semesters (e.g. 1 = ``Fall 2014'') in order to allow the model to learn relative temporal importance which is transferrable to future semesters.

\subsubsection{Encoder}
The Transformer encoder consists of a stack of identical layers, each performing two fundamental operations. First, it analyzes relationships between sequence elements through multi-head self-attention, then processes this information via a position-wise feed-forward network. Both operations employ residual connections followed by layer normalization to facilitate training \cite{ba2016layer}.

The multi-head attention mechanism enables the model to simultaneously examine representations from different viewpoints—analogous to multiple analysts focusing on different aspects of the same text:
\begin{align*}
\text{MultiHead}(Q, K, V) = \text{Concat}(\text{head}_1, \ldots, \text{head}_h)W^O
\end{align*}

Each attention head operates independently, calculating relevance scores between sequence elements:
\begin{align*}
\text{head}_i = \text{Attention}(QW_i^Q, KW_i^K, VW_i^V)
\end{align*}

The attention function uses scaled dot-product attention, computing similarity scores between queries and keys to create weighted combinations of values:
\begin{align*}
\text{Attention}(Q, K, V) = \text{softmax}\left(\frac{QK^T}{\sqrt{d_k}}\right)V
\end{align*}

The scaling factor $\sqrt{d_k}$ prevents gradient saturation by keeping attention scores within an optimal range for the softmax function \cite{vaswani2017attention}.

After capturing contextual relationships, the feed-forward network processes each position's representation individually:
\begin{align*}
\text{FFN}(x) = \max(0, xW_1 + b_1)W_2 + b_2
\end{align*}

This dual-stage processing—relationship analysis through attention followed by individual transformation through feed-forward networks—enables the model to capture dependencies regardless of sequential distance, a significant advantage over traditional recurrent architectures \cite{vaswani2017attention}.

In our context, the encoder attempts to capture the complex course-grade-major relationships and model them for later use in the model. That is, by representing the course and grade history as a latent vector in high-dimensional space, it can model these relationships in a way that is highly flexible, while also not needing to rely on human intervention (e.g. marking one course as a prerequisite for another, marking a course as being a senior-level course, etc).

\subsubsection{Decoder}
The decoder follows a similar structure to the encoder but includes an additional sub-layer that performs multi-head attention over the output of the encoder. Each decoder layer contains three sub-layers: a masked multi-head self-attention mechanism, a multi-head cross-attention mechanism that attends to the encoder output, and a position-wise feed-forward network.

For the purposes of course and grade prediction, the decoder layer reads in the latent vector created by the encoder layer, and utilizes a multi-head attention mechanism to attend to past courses and grades. The model learns the relevance of course history on grade and course prediction through the training process.

\subsubsection{Prediction Layer}
After passing through the Transformer decoder, a linear layer is then used to convert the outputs to the proper size. In our case that corresponds to size $2\cdot N_{course}$, where $N_{course}$ is the total number of possible courses (including the \texttt{<OTHER>} token). We double this because we predict both the course and the corresponding grade for the course. Note that we \textit{do not} predict the major, as this is already included in the input sequence, thus the output sequence is one term shorter than the input.

\subsubsection{Log-Softmax Layer}
We utilize the Kullback-Liebler loss function to measure course prediction, as discussed below. This loss function expects the predictions to be in log-space to avoid underflow issues. Therefore we pass logits through the log-softmax layer which computes

\begin{equation*}
    \text{LogSoftmax}(c_i) = \log \biggl( \frac{\exp(c_i)}{\sum_j \exp(c_j)} \biggr)
\end{equation*}

for all courses $c_i$.

\subsection{Training}
We group our data into batches of size 32, and train on a single NVIDIA RTX 3090 Ti for 20 epochs. We utilize the AdamW optimizer \cite{loshchilov2017decoupled}, along with a cosine annealing warm restart learning rate scheduler \cite{loshchilov2016sgdr} with $T_0 = T_{\text{mult}} = 5$.

\subsubsection{Loss Function}
Typically, sequential models are evaluated one token at a time, using either mean squared error (MSE) loss for regression tasks or cross-entropy loss for classification tasks. However, our task is unique in that it involves both regression (grades) and classification (courses). In addition, our classification task, while modeled as a sequence of courses, in fact has multiple tokens occurring simultaneously. This is because students typically take multiple courses in a semester, and there is no natural ordering for these courses. Therefore, a token-by-token evaluation using something like cross-entropy could potentially misclassify predictions as being incorrect, simply because the order predicted did not match the order listed. For example, if the student's target courses were listed as $(c_1, c_2, c_3)$, but the model predicted $(c_2, c_3, c_1)$, then evaluating this token-by-token would show zero correct predictions, which is undesirable. We solve this problem by viewing the courses taken in a given semester as a probability distribution over all possible courses.

We introduce a novel loss function combining KL-divergence over the predicted distribution of courses and MSE for grade regression, designed specifically for set-based prediction. Traditional cross-entropy-based sequence losses are inadequate for unordered output sets such as semester course baskets. Our loss formulation is inspired by general multiset prediction frameworks in ML but has not been previously applied in education data mining \cite{welleck2018loss}. Alternative set-based losses such as Hungarian matching \cite{kuhn1955hungarian} require solving an assignment problem at each training step, adding computational overhead. Chamfer loss \cite{wu2021density}, while permutation-invariant, treats predictions as point sets without probabilistic interpretation. Our KL-based formulation naturally handles the variable-size course sets while providing interpretable probability distributions over the course catalog.

We treat all courses within a semester as occurring simultaneously utilizing positional encoding, and thus the target is a binary vector corresponding to a one-hot encoding of the courses. We then normalize this so that the entries sum to one by dividing by the total number of courses taken in the semester. This way we can view the target as a probability distribution $P$ over all possible courses. The prediction is also a probability distribution $Q$ over all possible courses. We then use the Kullback–Leibler divergence (KL divergence) to evaluate the loss:

\begin{equation*}
    \displaystyle D_{KL}(P\ \lVert\ Q) = \sum_{i \in I}P(c_i)\log\biggl( \frac{P(c_i)}{Q(c_i)} \biggr).
\end{equation*}

Here $I$ is the indices for all possible courses, and $c_i$ indicates the course in the $i$'th position.

We note that course selection can be viewed more simply as a multi-label classification problem, where, given the course history up to that point, the task is simply to predict the next course and grade. Viewed this way, a natural loss function for the course prediction would be binary cross entropy, according to whether or not the correct course is predicted. However, using the KL divergence approach discussed here have several benefits. First, it naturally encodes course load in a given semester. Doing so aligns with the literature on academic momentum discussed in the introduction. In addition, it allows a natural modeling of the interdependency of concurrently taken courses. For example, students taking multiple difficult courses in the same semester may struggle, and viewing course load as a distribution over courses naturally addresses this. Finally, we ran an ablation study comparing the above loss function to binary cross entry for course prediction, and noted grade prediction within 1-2\% of each other between the two models for both MAE and MSE. Therefore, while we believe that the KL divergence-based approach has learning analytics-aligned benefits with no prediction drawback, these results suggest the option of utilizing a BCE loss as as reasonable alternative.

For GPA we calculate the MSE loss for the predicted grade against the actual grade. We mask the padding values so that the loss is not calculated for padding courses or for start/end of sequence tokens. We also mask grade predictions for courses not taken, since we are only interested in measuring prediction quality when we have a ground truth to compare the prediction against.

Finally, we balance the relative importance of each of these two loss values (course and grade) through the use of a weighting factor. That is

\begin{equation}
    \text{Loss}_{\text{total}} = \text{Loss}_{\text{course}} + \omega \cdot\text{Loss}_{\text{grade}}
\end{equation}

where $\omega \in \mathbb{R}$ is fixed. Note that $D_{KL}$ is positive but unbounded, and MAE is unbounded and depends on the scale of the grades. Therefore, the precise value of $\omega$ will depend heavily on the data, and should be optimized using cross validation during training. Through our experiments we found that a value of $\omega$ near one was optimal. 

\subsubsection{One-Step-Ahead Training}
While courses within a semester are taken simultaneously, Transformer models still make predictions sequentially. That is, at any given step the set of courses, grades and semesters, ignoring padding, may look like

\begin{align}
    c &= (\texttt{<SOS>}, c_1, c_2, \ldots, c_n, \texttt{<EOS>}) \nonumber \\
    g &= (0., g_1, g_2, \ldots, g_n, 0.) \nonumber \\
    t &= (t_{\texttt{<SOS>}}, t_1, t_2, \ldots, t_n, t_{\texttt{<EOS>}})
\end{align}

where $c_i$ indicates a course, $g_i$ indicates the grade in course $c_i$, and $t_i$ indicates the semester in which course $c_i$ was taken. However, during inference time, we only know information up to a certain point, and wish to predict future courses and grades. To format this into a sequence learning problem, we utilize one-step-ahead prediction, which truncates each sequence up to position $i \geq 1$, and using the first $i$ entries as the input and the remaining $n - i$ entries as the target. For example, with $i=2$ we have

\begin{align}
    c_{\text{in}} &= (\texttt{<SOS>}, c_1) &  c_{\text{out}} &= (c_2, \ldots, c_n, \texttt{<EOS>}) \nonumber \\
    g_{\text{in}} &= (0., g_1) &  g_{\text{out}} &= (g_2, \ldots, g_n, 0.) \nonumber \\
    t_{\text{in}} &= (t_{\texttt{<SOS>}}, t_1) &  t_{\text{out}} &= (t_2, \ldots, t_n, t_{\texttt{<EOS>}})
\end{align}

The model then iterates through all values of $i$ from 1 to $n$.

\section{Results and Discussion}\label{sec4}
\subsection{Comparison Methods}
In order to understand both the performance of our model but also the relative importance of various features and architecture choices, we compare our model to the following alternatives and variants:

\begin{itemize}  
    \item \textbf{Transformer without course predictions (``OnlyGradesTransformer''):} During training we predict the grades in the courses actually taken, but not the courses taken. 
    \item \textbf{Encoder-Decoder LSTM (``EncDecLSTM''):} We utilize an encoder-decoder LSTM with attention. This closely mirrors our Transformer model setup, but replaces the Transformers in the encoder and decoder with unidirectional LSTMs with attention.
    \item \textbf{Unidirectional LSTM (``UniLSTM''):} We also implement a unidirectional LSTM without the encoder-decoder architecture described above. This LSTM also does not use any attention mechanism, and represents a traditional LSTM architecture.
    \item \textbf{Graph Neural Network (``GNN''):} The model operates on a bipartite graph where students and courses constitute two distinct node types, connected by enrollment edges. For each prediction task (predicting a student's courses and grades for semester $t$), we construct a subgraph containing the target student's enrollment history from semesters $1$ through $t-1$, along with enrollments from collaborating students who share at least one course with the target student. This collaborative neighborhood provides the GNN with signals about how similar students performed in related courses.
    \item \textbf{XGBoost (``XGBoost''):} We treat our data as tabular with rows corresponding to a particular (student, semester, course) tuple, where we encode all relevant data (student ID, semester, course taken, grade in course) as columns with no explicit sequential connections. Note that the inclusion of semester means the model can still model the concurrency of courses taken in the same semester. Therefore, this approach allows us to closely compare the performance of a powerful machine learning algorithm against a model deep learning approach such as an LSTM or Transformer.
\end{itemize}

For each model we evaluate grade prediction using MSE and mean-absolute error (MAE), and compare the results to the model proposed in this paper (``TRACE''). We conducted cross-validation hyperparameter tuning for the XGBoost model to determine optimal hyperparameters. The results are shown in Table \ref{tab:results}.

Because of the relative novelty of graph neural networks, in particular in the use of learning analytics, we provide a brief overview of our architecture and design choices.

\subsubsection{Graph Neural Network Architecture}
Our GNN operates on a bipartite graph where students and courses constitute two distinct node types, connected by enrollment edges. Predictions, both of courses and grades for semester $t$, are made by constructing a subgraph containing the target student's enrollment history from semesters $1$ through $t-1$, along with enrollments from collaborating students who share at least one course with the target student.

Each enrollment edge carries two features: the normalized semester rank (indicating when the course was taken relative to the student's academic career) and the normalized grade received. These edge features serve an analogous role to the positional encodings in TRACE, encoding temporal information about when each course-grade observation occurred.

Student node features are computed as the concatenation of two components: (1) a learned embedding of the student's major, and (2) the mean of learned embeddings for all courses in the student's history. Critically, we do not use learnable per-student parameters, ensuring the model can generalize to unseen students and maintaining parity with TRACE, which similarly lacks student-specific parameters. This aggregation-based initialization allows the model to represent students based on their academic trajectory rather than memorized identities.

Course node features are initialized from a learned embedding table, identical to the course embeddings used in TRACE. Both student and course features are projected to a common dimensionality of 128 to match TRACE's hidden dimension.

The core of the GNN consists of two layers of edge-conditioned bipartite message passing. Each layer performs bidirectional communication: first from students to courses, then from courses to students. The message passing operation computes messages as $m_{ij} = \text{MLP}([h_i | e_{ij}])$, where $h_i$ is the source node's features and $e_{ij}$ is the encoded edge features. Messages are aggregated at target nodes via mean pooling, then combined with the target node's current features through an update MLP. Residual connections and layer normalization stabilize training. We experimented with three, four or five layers for longer distance message passing, but found no improvement in test set performance, at significantly higher computational cost, and with evidence of overfitting.

After two rounds of message passing, each student node has aggregated information from their enrolled courses, and transitively from other students who took those same courses. This two-hop receptive field allows the model to capture collaborative filtering signals—students who took similar courses will develop similar representations.

For the final prediction, we extract the target student's updated embedding and score it against all possible courses. For each course, we concatenate the student embedding with the course embedding and pass it through two separate prediction heads: a course prediction head outputting a binary logit (trained with binary cross-entropy loss and negative sampling), and a grade prediction head outputting a value in $[0, 1]$ via sigmoid activation (trained with MSE loss). This dual-head architecture mirrors TRACE's joint prediction of courses and grades.

We note that the GNN model required roughly 50\% more training time than TRACE on identical hardware, likely due to the computational overhead of constructing per-prediction subgraphs and message passing across the student-course bipartite structure. This suggests that TRACE offers favorable accuracy-to-compute tradeoffs for institutions with limited computational resources.

\subsection{Results}
For each of the model and data setups described above, we train the model on the data over twenty epochs. For all models we utilize MSE loss for grade predictions. For models which also predict courses we utilize K-L divergence loss for course predictions. After training we then test on a held-out test set consisting of 10\% of all students. We evaluate grade predictions using mean absolute error, in order to understand the error on a $[0, 1]$-scaled GPA scale, as well as MSE. Grade metrics are computed over the student's actual enrolled courses. That is, we extract the model's predicted grade for each course the student actually took, so course prediction accuracy does not influence the reported grade error. This is the natural approach, as we do not have grade data for courses students didn't actually take.

We see from Table \ref{tab:results} that the Transformer model described in this paper drastically outperforms the XGBoost model. The improvement over the XGBoost model is largely because tree-based models are not naturally suited to sequential learning tasks.

Compared to the two LSTM models, the Transformer model shows a significant improvement, with a MSE approximately 30\% lower than the LSTM models, and a MAE approximately 15\% lower. We note that the encoder-decoder LSTM uses an attention mechanism, whereas the unidirectional LSTM does not. Regardless, both models have similar performance. This indicates that the key bottleneck is not the attention mechanism, but rather the underlying model architecture. By design, LSTMs process data sequentially, updating their hidden state after each token processed. The Transformer, on the other hand, allows every token to attend to every other (non-masked token). This is likely very important in the context of grade prediction, because relevant courses may have been taken several semester prior, and multiple different courses may all contribute to the prediction in the current course. 

Compared to the graph neural network, TRACE has an MSE/MAE 15\%-20\% lower than the graph neural network. The GNN assumes locality in the graph, namely that a student's performance is most influenced by their immediate course neighborhood and by similar students who share that neighborhood. The Transformer model makes no such locality assumption, instead learning which historical observations are relevant through attention. For academic data where prerequisite chains and course sequences matter, the Transformer's ability to attend to distant but relevant courses may prove advantageous.

Finally, comparing the two Transformer-based models we note a stark difference in performance. The TRACE model achieved a mean absolute error of 0.1339, a 46.4\% reduction compared to the `OnlyGradesTransformer`'s MAE of 0.2496, and a mean squared error 3.5 times lower. This large difference demonstrates that the Transformer model architecture on its own is not sufficient to guarantee strong performance. A key benefit of the Transformer architecture is the robust latent representation it learns, and if these are not leveraged in the case of courses, then predictions suffer greatly.

\begin{table}[h]
\centering
\begin{tabular}{|l|l|l|}
\hline
\multicolumn{1}{|c|}{\textbf{Model}} & \multicolumn{1}{c|}{\textbf{MSE}} & \multicolumn{1}{c|}{\textbf{MAE}} \\ \hline
TRACE                                 &          \textbf{0.0392}          &   \textbf{0.1339}\\ \hline
OnlyGradesTransformer                &              0.1400               &   0.2496         \\ \hline
EncDecLSTM                           &              0.0530               &   0.1608         \\ \hline
UniLSTM                              &              0.0544               &   0.1619         \\ \hline
GNN                              &             0.0495               &   0.1640         \\ 
\hline
XGBoost                              &              1.3317               &   0.8818         \\ \hline
\end{tabular}
\caption{Overview of results comparing TRACE with baseline models.}
\label{tab:results}
\end{table}

As discussed above, the primary purpose of the model is grade prediction, and course prediction is utilized to encourage the model to learn meaningful course vector representations. To assess the quality of course predictions, we computed standard classification metrics for the TRACE model on the test set ($P=0.5076$, $R=0.4090$, $F1=0.4432$).

These metrics reflect several challenges inherent to course prediction. First, with 360 possible courses in the catalog, the task is substantially more difficult than typical binary or small-cardinality classification problems. Second, many enrollment decisions depend on factors outside the model's scope, including student preferences, scheduling constraints, advisor recommendations, 
and choice among equivalent courses (e.g., different courses satisfying the same general education requirement).

As a baseline, random selection of 5 courses from the 360-course catalog would yield an expected precision of approximately 0.014 (5/360), highlighting that even modest metrics represent substantial predictive signal.

Critically, these course prediction metrics demonstrate that the auxiliary task successfully regularizes the learned representations: the model achieves a 46\% reduction in grade prediction MAE compared to the OnlyGradesTransformer baseline (Table \ref{tab:results}), confirming that imperfect but informative course predictions substantially improve the primary task.

\subsubsection{Feature Ablation Analysis}
To assess the contribution of individual features, we trained variants of TRACE with different information removed. As discussed above, the `OnlyGradeTransformer` model keeps courses as input, but removes them from prediction. As seem in Table \ref{tab:results}, this resulted in a significant drop in performance.

To assess the contribution of student major, we trained a variant of TRACE with major removed from the input. This ablation resulted in marginally improved performance (MSE: 0.0360 vs. 0.0392; MAE: 0.1327 vs. 0.1339), suggesting that major provides no additional predictive signal beyond course history and may introduce slight redundancy.

We interpret this as evidence that a student's major is largely recoverable from their early course selections. A computer science student, for instance, often enroll in introductory programming and discrete mathematics within their first semesters. These patterns are likely sufficiently distinctive that the model learns major-like groupings implicitly. When major is provided explicitly, the model receives a redundant signal that marginally increases overfitting risk without contributing new information. However, the benefit of keeping an embedding of the major as the first entry in the sequence is that it shows how other student attribute data could be incorporated in a similar manner.

\section{Conclusion}\label{sec5}
A primary goal of this work is to provide a reproducible methodological framework that other institutions can adapt with their own data. The architecture requires no institution-specific feature engineering—course and major embeddings are learned from enrollment data alone, and the model can be retrained on any institution's historical records without structural modification. 

We introduced a novel Transformer-based architecture that jointly predicts both the set of courses a student will take and the grades they will achieve in them. This approach, coupled with a custom loss function, allows the model to learn the complex, non-sequential inter-dependencies between courses, major, and historical performance. We demonstrated that predicting courses taken, as well as grades, is fundamental in downstream tasks. Without prediction of courses, the model fails to learn robust course vector representations, which harms the primary task of grade prediction. Therefore, this work advances the body of learning analytics knowledge by demonstrating that for complex sequential prediction, modeling the structural properties of the sequence (i.e., concurrent semesters) and the inter-relationships of the outputs (i.e., joint course-grade prediction) can be as crucial as the choice of the underlying model architecture itself. 

While in our work the only non-course/grade data we had access to was student majors, our approach of prepending embeddings of student majors to the sequence gives a guide for how an institution could implement other student attribute data, such as demographics, high school data, etc. We release our training code to facilitate replication and extension. While single-institution validation is a limitation, our results establish a performance baseline and architectural template against which future cross-institutional studies can be compared.

\subsection{Principal Findings}
Our results demonstrate the superiority of this joint prediction approach, as well as the Transformer architecture. The model we developed significantly outperformed comparison models across key metrics. Specifically, our proposed model TRACE, which jointly predicts courses and grades, achieved a mean absolute error nearly half that of an identical Transformer architecture that only predicted grades, and a mean squared error 3.5 times lower. This confirms our hypothesis that forcing the model to learn meaningful course embeddings through the course prediction task provides a richer representation that directly benefits grade prediction accuracy.

Furthermore, when compared to a traditional sequence-to-sequence model using an Encoder-Decoder LSTM with attention, our Transformer model showed marked improvements. Regardless of whether an attention mechanism is utilized or not, and regardless of whether the LSTM is unidirectional or bidirectional, the Transformer model still solidly outperformed it. 

Despite their strengths, existing GNN-based approaches typically focus on predicting student outcomes (e.g., grades or pass/fail status) for a fixed course or time window, rather than modeling academic progression across semesters. Courses are usually treated as features or as sources of interaction edges, rather than as first-class prediction targets. Moreover, concurrency of course enrollment is handled implicitly through graph connectivity rather than explicitly as a structural unit of the prediction task. Temporal information, when included, is often incorporated through static snapshots or limited temporal aggregation rather than explicit sequence modeling.

\subsection{Early Alert Systems}
These findings pave the way for a new class of advising tools that move beyond simple risk-flagging to provide nuanced, data-driven insights into how specific course combinations may impact student success. For example, an early warning system could be constantly running in the background as student grade data is updated each semester. Results could be integrated into student advising. Our model achieves a grade prediction MAE of 0.0392 on a $[0, 1]$ GPA scale, which corresponds to 0.1568 on a typical $[0, 4]$ GPA scale. This means predictions are accurate, on average, to about half of the difference between a typical +/- grade (e.g. B vs B+). accurate enough to serve a meaningful purpose in an early alert system.

While models like this one don't replace human insight, they can help alert faculty and student support staff to key problems before they arise. It has been shown that even minor ``nudges'' can help students, and that early intervention is key \cite{smith2018improved} \cite{chen2014incentive} \cite{gordanier2019early} \cite{educsci16020233}. In spite of this, it is still common for universities to rely on manually entered instructor early alerts \cite{faulconer2013adoption} \cite{tampke2013developing}. Therefore, systems such as this one can help streamline this process.

\subsection{Limitations}
Despite these promising results, this study has several limitations. First, the model was trained and tested on data from a single, medium-sized private university. The student demographics, course offerings, and academic rigor at this institution may not be representative of other academic contexts, such as large public universities, community colleges, or institutions in other countries. The model's performance may not generalize without retraining on institution-specific data.

Second, our model's features were limited to student major, course history, and grade history. As noted in our introduction, a host of other factors, including socioeconomic status, student engagement, and non-cognitive skills, are known to influence academic success. The exclusion of these variables means our model captures only one dimension of the student experience. Historical grading data may reflect systemic biases related to instructor, demographic, or socioeconomic factors. Our model learns from these historical patterns and may therefore encode and propagate such biases in its predictions. Future work should incorporate fairness auditing to assess whether predictions exhibit disparate accuracy across student subgroups, and explore bias-mitigation strategies during training.

Finally, the model faces a "cold start" problem for new students with no academic history at the university. Similarly, newly introduced courses that are not in the training vocabulary are mapped to an \texttt{<OTHER>} token, which limits the model's predictive power for novel curricula until sufficient data is collected.

\subsection{Future Work}
The findings and limitations of this study suggest several avenues for future research. An immediate next step would be to validate the model's architecture and performance on datasets from a more diverse range of academic institutions to assess its generalizability.

Future work may also explore integrating permutation-invariant architectures such as Set Transformers \cite{lee2019set} or DeepSets \cite{zaheer2017deep}, which may better align with the unordered structure of semester-level course sets.

Future iterations could also aim to incorporate a richer feature set. Integrating data from Learning Management Systems (LMS), such as login frequency, forum participation, and assignment submission times, could provide a more granular view of student engagement. Demographic and financial aid data could also be included to model the impact of non-academic factors, provided that ethical and privacy considerations are carefully managed.

In terms of architecture choices, we compared a Transformer architecture to a graph neural network architecture. Both performed extremely well, especially as compared to common baselines. However, recent work has emerged in Graph Transformer models, which can ``automatically and implicitly learn and extract “meta paths” that are important for different downstream tasks'', which show promise for tasks such as those described in this paper \cite{hu2020heterogeneous}.

Lastly, exploring the practical application of the model's interpretability is a key future direction.

\section{Data Availability}\label{sec6}
The datasets generated and/or analyzed during the current study are not publicly available due to the data originating from private data from a private university. However, an anonymized subset of the data is available at \url{https://github.com/paulsavala/TRACE-grade-prediction}.


\backmatter

\section{Declarations}

\subsection*{Funding}
No funding was received for this work.

\subsection*{Ethics approval and consent to participate}
Not applicable.

\subsection*{Consent for publication}
Not applicable.

\subsection*{Materials availability}
The datasets generated and/or analyzed during the current study are not publicly available due to the data originating from private data from a private university. However, an anonymized subset of the data may be available in an anonymized form from the corresponding author on reasonable request.

\subsection*{Code availability}
Code is available at \url{https://github.com/paulsavala/TRACE-grade-prediction}.

\subsection*{Acknowledgments}
We would like to acknowledge the help of St. Edward's University in sourcing this data, and Dean Jon Hodge for supporting this work.

\bibliography{sn-bibliography}

\end{document}